%% file: main.tex
\documentclass[runningheads]{llncs}
\usepackage{graphicx}
\usepackage{hyperref}
\usepackage{amsmath}
\usepackage{amssymb}
\usepackage{mathtools}
\usepackage{subcaption}
\usepackage{booktabs}
\usepackage[misc,geometry]{ifsym}
\usepackage{threeparttable}
\usepackage{svg}
\usepackage[english]{babel} % English language/hyphenation
\usepackage[babel]{csquotes} % To use \enquote
\usepackage{hyperref} % To use \url
\definecolor{Blue}{RGB}{0,0,255}
\definecolor{Purple}{RGB}{230,230,250}
\definecolor{Red}{RGB}{255,0,0}

\usepackage[framemethod=tikz]{mdframed} % To add block-comments
\usepackage{float}
\begin{document}

\title{RoFormer for Position Aware Multiple Instance Learning in Whole Slide Image Classification}

\titlerunning{RoFormer encoder for MIL}

\author{Etienne Pochet \and Rami Maroun \and Roger $\text{Trullo}^\text{\Letter}$}
\institute{Sanofi, Chilly Mazarin, France
\\ \email{etienne.pochet@sanofi.com}
\\ \email{rami.maroun@sanofi.com}
\\ \email{roger.trullo@sanofi.com}}

\maketitle

\renewcommand{\sectionautorefname}{\S}
\renewcommand{\subsectionautorefname}{\S}

\input{0_abstract}

\section{Introduction}
\label{introduction}
\input{1_introduction}

\section{Related Work}
\label{related}
\input{2_related_works}

\section{Methods}
\label{Methods}
\input{3_methods}

\section{Experiments}
\label{Experiments}
\input{4_experiments}

\section{Conclusion}
\label{conclusion}
\input{5_conclusion}

\bibliographystyle{splncs04}
\bibliography{mil_bib}

\end{document}

%% file: 0_abstract.tex
\begin{abstract}
Whole slide image (WSI) classification is a critical task in computational pathology. However, the gigapixel-size of such images remains a major challenge for the current state of deep-learning. Current methods rely on multiple-instance learning (MIL) models with frozen feature extractors. Given the the high number of instances in each image, MIL methods have long assumed independence and permutation-invariance of patches, disregarding the  tissue structure and correlation between patches. Recent works started studying this correlation between instances but the computational workload of such a high number of tokens remained a limiting factor. In particular, relative position of patches remains unaddressed.

We propose to apply a straightforward encoding module, namely a RoFormer layer , relying on memory-efficient exact self-attention and relative positional encoding. This module can perform full self-attention with relative position encoding on patches of large and arbitrary shaped WSIs, solving the need for correlation between instances and spatial modeling of tissues. We demonstrate that our method outperforms state-of-the-art MIL models on three commonly used public datasets (TCGA-NSCLC, BRACS and Camelyon16)) on weakly supervised classification tasks. 
\\
\\
Code is available at \href{https://github.com/Sanofi-Public/DDS-RoFormerMIL}{https://github.com/Sanofi-Public/DDS-RoFormerMIL}

\end{abstract}

%% file: 1_introduction.tex
Whole slide images (WSI), representing tissue slides as giga-pixel images, have been extensively used in computational pathology, and represent a promising application for modern deep learning \cite{gadermayr2022multiple}\cite{madabhushi2009digital}\cite{shamshad2023transformers}. However, their size induces great challenges, both in computational cost for the image processing and lack of local annotations. Hence, the WSI classification has been studied as a weakly-supervised learning task, treating the image as a bag of instances with one label for the entire bag.  Multiple instance learning (MIL) techniques have had success in this scenario \cite{ilse2018attention}\cite{li2021dual}\cite{lu2021data}. WSIs are first divided into patches, then a feature extractor generates features for each of them and finally the feature vectors are aggregated using global pooling operators to get an image-level prediction. 

\begin{figure}[H]
\centering
\includegraphics[width=1.0\linewidth]{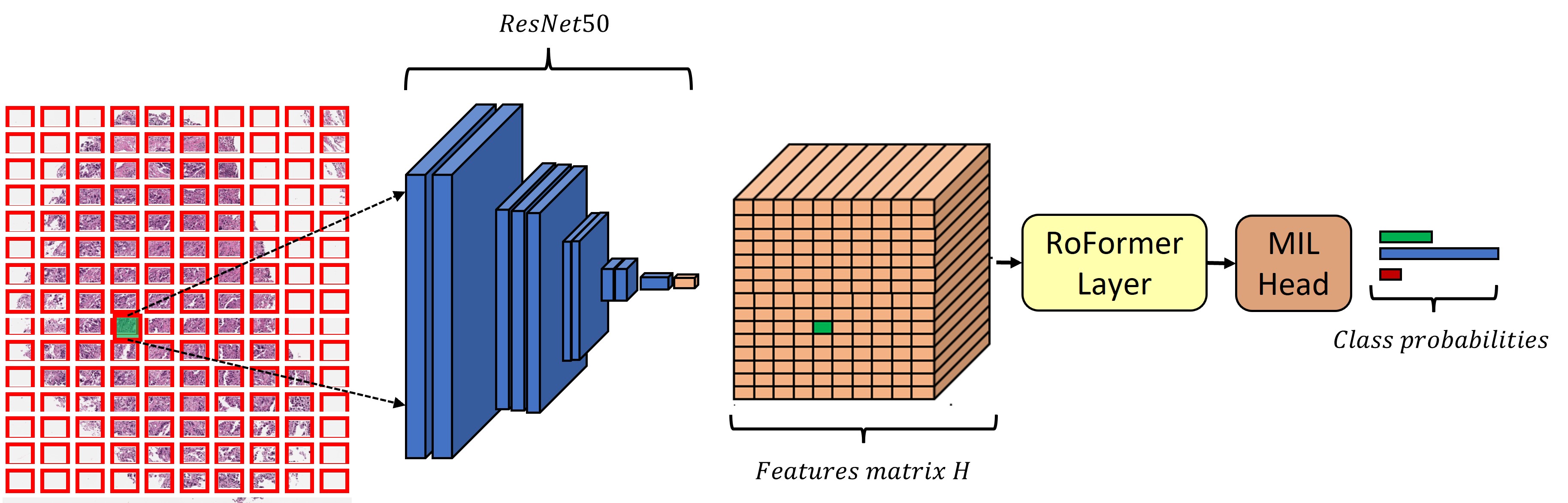}
\caption{Model Pipeline}
\label{fig:pipeline}
\centering
\end{figure}

Given the high number of instances in each giga-pixel image, works like ABMIL\cite{ilse2018attention} or CLAM\cite{lu2021data} assumed independence between patches and applied permutation-invariant methods to avoid comparing patches, reducing the computational workload. These assumptions allow to treat large number of instances efficiently but prevent from modelling the correlation between patches and the tissue spatial structure.

More recent studies attempted to bridge these gaps. Works like HIPT\cite{hipt} or HAG-MIL\cite{hag-mil} use hierarchical transformers to aggregate local and global information. Closer to our work, self-attention has been used as an encoding layer, without positional-encoding to capture dependencies between patches \cite{rymarczyk2021kernel}\cite{myronenko2021accounting}. TransMIL\cite{transmil} reshapes the image into a square feature map to apply group convolution and Nystrom method \cite{nystromformer} which is just an approximation to a full self attention, in an attempt to encode spatial information into the patch features. 

In light of the recent advances in transformers research, we propose a straightforward encoding module relying on a RoFormer encoder layer; a transformer block with rotary position embedding (RoPE)\cite{roformer}. The proposed module, implemented with memory-efficient attention \cite{memefficient}, can perform full self-attention without resorting to any approximation with relative position encoding for large WSIs, with arbitrary and irregular shapes, on consumer GPUs.
We use the coordinates of patches in the slide to apply the relative position encoding, treating the WSI as a 2D sequence of instances rather than an unordered bag. 
This encoder is agnostic to the downstream model and we experiment with both ABMIL\cite{ilse2018attention} or DSMIL\cite{li2021dual}.
Our model eventually fits in 8GB GPUs.  Figure \ref{fig:pipeline} summarizes our approach. 

We evaluate the influence of this encoding module alongside proven MIL models on three of the main public benchmarks for H\&E stain WSI classification: TCGA-NSCLC\cite{tcga}, BRACS\cite{bracs} and Camelyon16\cite{camelyon} and show a consistent improvement on classification performance with our proposed method.

%% file: 2_related_works.tex
\subsection{Multiple Instance Learning}
\label{MIL}
MIL in WSIs classification has had great success with using a two-step modeling where
each WSI is considered as a bag and its patches are the instances.
A frozen feature extractor generates a feature vector for each instance. Then a pooling operation generates the image-level prediction. The attention-based pooling introduced in ABMIL, widely adopted in later works \cite{lu2021data}\cite{hipt}\cite{yao2020whole}, introduces a learnable class token which is compared in global attention to all patches features. The slide representation $z$ is computed as :
\begin{equation}
z =\sum_i \alpha_{i} H_i 
\end{equation}

\begin{equation}
\alpha_{i} = (\text{softmax}\left(CH^T\right)H)_i
\end{equation}
with $H$ the feature matrix of all patches, $C$ the learnable class token. Linear projection and multi-head formulation are omitted in the equation for simplicity.

DSMIL \cite{li2021dual} modified this pooling function to perform global attention of a critical patch, identified with a learned classifier, over all other patches. 

Both formulations enable  all patches to influence the final prediction with limited computations. However, they do not model the dependencies between patches nor their relative positions. The ability to encode patch representations with respect to other patches and tissue structures remains a challenge.

\subsection{Attempts at position-aware MIL}
\label{transmil}
 Among notable attempts to model spatial dependencies between patches, the two most related to our work are, to the best of our knowledge, HIPT and TransMIL. 

\paragraph{HIPT} In this work, patches are aggregated into macro regions of $4k$*$4k$ pixels, instead of the 256*256 original patches, during feature extraction. A multilayer ViT\cite{vit} then encodes regions features before feeding them to the global pooling layer. This encoder, trained at the same time as the attention pooling, is then exposed to a much smaller number of tokens and can perform full self-attention with absolute position encoding. The interest of aggregating patches into larger regions is out of the scope of this work and we will investigate the use of a similar ViT in the setup of small patches.

\paragraph{TransMIL} This work leverages the inductive bias of CNNs to aggregate context information, after squaring the $N$ patches into a $\sqrt{N}*\sqrt{N}$ feature map, regardless of the initial tissue shape. Dependencies between instances are also modeled using a Nystrom-approximation attention layer\cite{nystromformer}.

\subsection{Relative position encoding}
Position encoding in long sequence transformers is still an open question, even more so for WSI where the tokens are irregularly located. Absolute position encoding, of ViTs, helps modeling dependencies between instances \cite{hipt} but  the arbitrary shapes, sizes and orientations of WSIs may require relative position encoding. However, methods with discrete number of relative positions \cite{rpe}\cite{relativepevit} would not account for the large irregular shapes of tissues in WSIs. Rotary Position Encoding (RoPE) \cite{roformer} provides an elegant solution to this problem by modifying directly the key-query inner product of self-attention, with minimal  computation overhead  and handling any maximum size of relative position, which matches our very long sequence setup.

%% file: 3_methods.tex
\begin{figure}[H]

\begin{subfigure}{.45\textwidth}
  \centering
  \includegraphics[width=\linewidth]{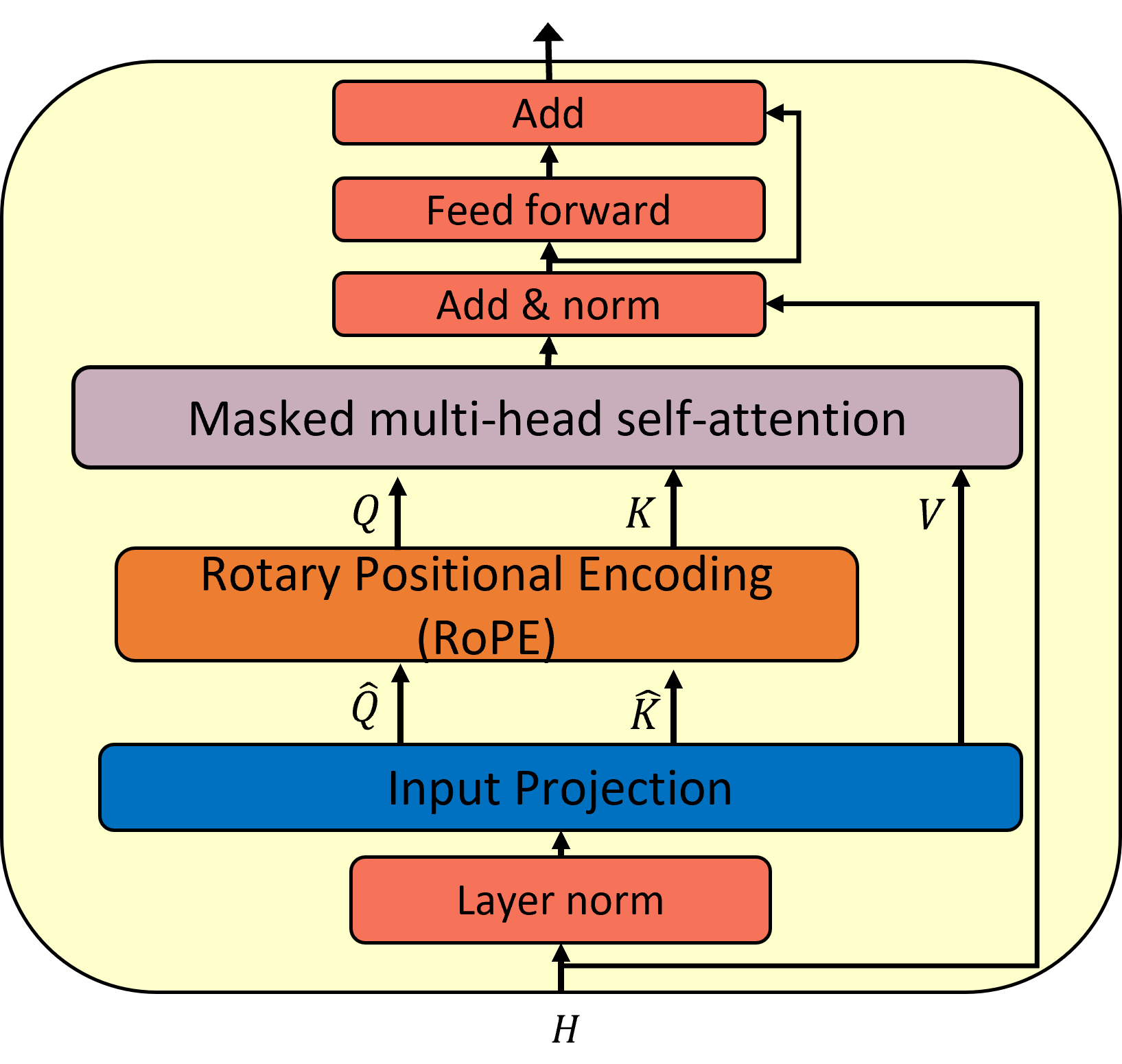}
  \caption{Roformer Layer}
  \label{fig:roformer}
\end{subfigure}%
\begin{subfigure}{.45\textwidth}
  \centering
  \includegraphics[width=0.55\linewidth]{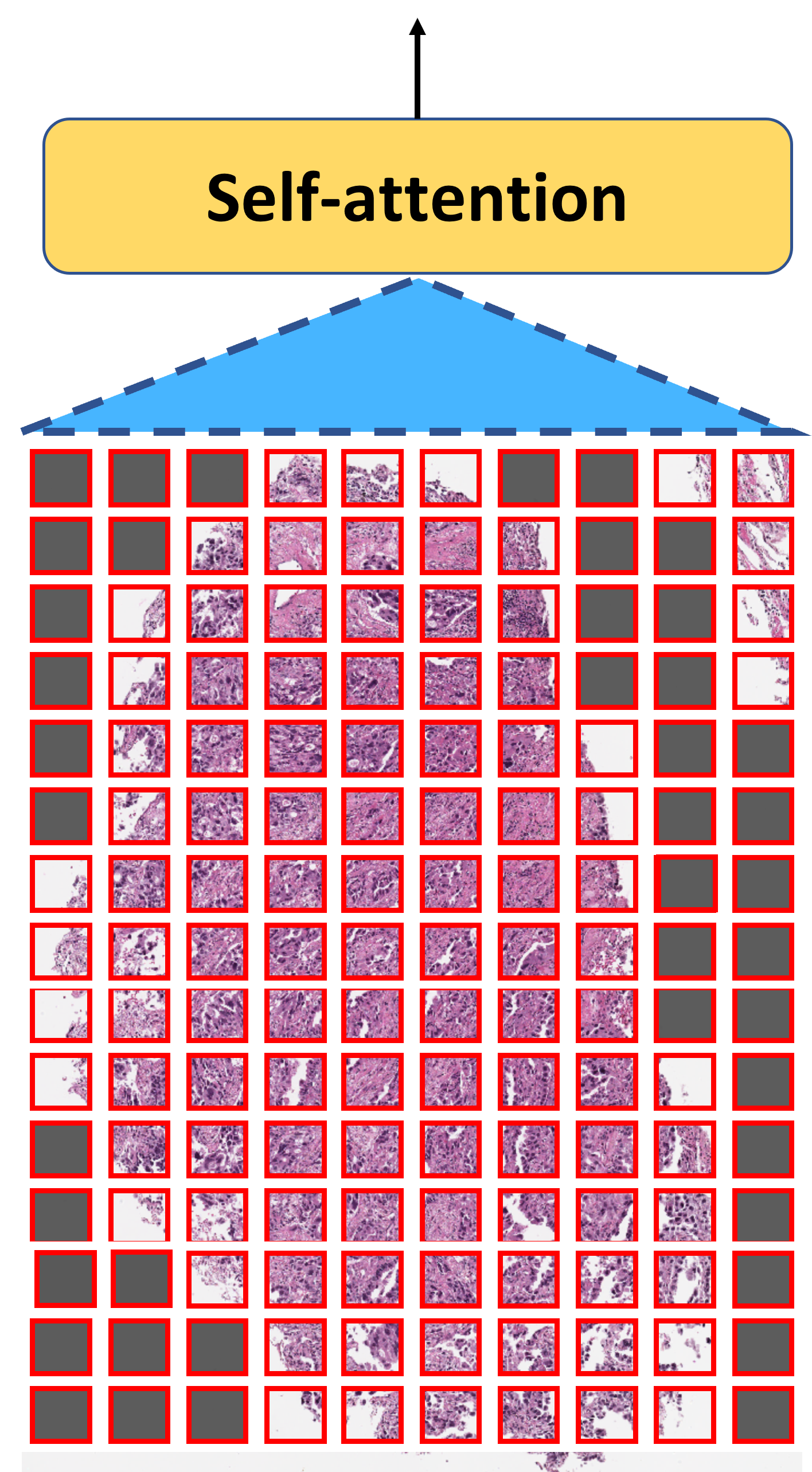}
  \caption{Masked MultiHead Self Attention}
  \label{fig:mha}
\end{subfigure}
\caption{Key components of the RoFormer encoder}
\label{roformer}
\end{figure}

In this section, we describe the encoding module we use, and the required modifications to the WSI patching. The feature extraction, with a frozen encoder, as well as the global attention pooling layer are out of the scope of this work. The approach is summarized in Fig. \ref{fig:pipeline}. The coordinates and mask positions of each embedded patch are kept and used in a RoFormer layer with masked self-attention (Fig. \ref{roformer}). Encoded outputs are fed to an MIL classification head.

Unless stated otherwise, we applied the same feature extraction as described in CLAM\cite{lu2021data}.

\subsection{Patching}
As it is typically done, the WSI is divided into small patches, of size $P*P$ pixels, using only the tissue regions and disregarding the background.
However, instead of treating the patches independently from their location in the slide, we keep the coordinates of each patch. In particular, we consider the WSI as a grid with squares of size $P$ and locate each patches on this grid. 
Simply put, we compute for each patch: 

\begin{equation}
\label{grid}
(x_{grid}, y_{grid}) = (\left \lfloor{\frac{x_{pixel}}{P}}\right \rfloor, \left \lfloor{\frac{y_{pixel}}{P}}\right \rfloor ) \end{equation}
with $x_{pixel}$ and $y_{pixel}$ the pixel coordinates of the patch in the WSI.

The patches, and their corresponding feature vectors, can now be seen as tokens in a 2D sequence, as in ViT. Because we only consider patches containing tissue and disregard the background, the grid can be sparse and of arbitrary size, depending on the tissue represented in the WSI. Patches representing background are masked out for the following steps.

\subsection{Self-attention encoding}
The key component of the encoding module is a simple RoFormer layer (Fig \ref{fig:roformer}). The full self-attention mechanism allows for each patch to update its feature vectors with respect to all other patches in the slide, effectively modeling the interactions between instances and tissue regions. The relative position encoding introduces spatial information and models the tissue structure. Masked self-attention is applied to discard the background patches (Fig \ref{fig:mha}).

\subsubsection{Efficient self-attention}
The $O(n^2)$ memory and time complexity of self-attention, with respect to $n$ the number of patches, has prevented its use in WSI classification as digital slides may contain tens of thousands of patches.  

However, recent breakthrough implementations such as Flash attention\cite{flashattention} or Memory Efficient attention\cite{memefficient}, considerably reduced the burden of those operations while still providing exact full self attention. We used memory-efficient attention for both the encoding self-attention and the global-attention pooling.

\input{rotary_draft}

%% file: rotary_draft.tex
\subsubsection{Rotary position encoding}
As discussed in section 
\ref{related}, relative position encoding for WSI with large numbers of tokens and irregular locations is challenging. To address this issue, we leverage RoPE to encode position of the patches. The relative nature, translation-invariance and ability to handle any number of relative positions of this method enables to encode spatial dependencies in the WSI setup.

RoPE is originally defined, for 1D text sequences, as: 
\begin{equation}
RoPE(h, m) = \begin{pmatrix} h_1 \\ h_2 \\ \vdots \\ h_{d-1}\\ h_d  \end{pmatrix} 
\otimes
\begin{pmatrix} \text{cos } m\theta_1 \\ \text{cos } m\theta_1\\ \vdots \\ \text{cos } m\theta_{d/2}\\ \text{cos } m\theta_{d/2}\end{pmatrix}
+
\begin{pmatrix} - h_2 \\ h_1 \\ \vdots \\ - h_{d}\\ h_{d-1}  \end{pmatrix} 
\otimes
\begin{pmatrix} \text{sin } m\theta_1 \\ \text{sin } m\theta_1\\ \vdots \\ \text{sin } m\theta_{d/2}\\ \text{sin } m\theta_{d/2}\end{pmatrix}
\end{equation}
with ${h \in \mathbb{R}^d}$ the feature vector, $m$ the index of the token in the input sequence and $\theta_i = 10000^{-2i/d}$.

We apply RoPE to our 2D coordinates system, using the same extension as ViT for sin-cos absolute position encoding. From the input feature vector $h \in\mathbb{R}^d$ with coordinates $(x_{grid}, y_{grid})$, $\frac{d}{2}$ features are embedded with coordinate $x_{grid}$, the other $\frac{d}{2}$  with $y_{grid}$ and the results are concatenated:
\begin{equation}
RoPE_{2D}(h, (x_{grid}, y_{grid})) = [RoPE(h_{[1:d/2]}, x_{grid}), RoPE(h_{[d/2:d]}, y_{grid})]
\end{equation}
This embedding is applied to each patch in the key-query product of the self-attention layer.

%% file: 4_experiments.tex
We evaluate the effect of our position encoding module on three standard WSI classification datasets and evaluate the influence of our design choices in an ablation study.

\subsubsection{Implementation details}
\label{implem}
For all three datasets, we used $256*256$ patches at a magnification of x20, downsampling from x40 when necessary, and ResNet50-ImageNet \cite{resnet} as feature extractor. We train the MIL models with a Adam optimizer, learning rate of $0.0001$, batch size of 4 and a maximum of 50 epochs with early stopping. We implemented our code using Pytorch and run it on a single 16 GB V100 GPU. We use the xFormers implementation \cite{xFormers2022} for memory-efficient attention.

\subsubsection{Datasets} \label{datasets} We leverage three public datasets of H\&E slides that effectively cover the cases of binary, multi-class and imbalanced classifications.

First, we used 1041 lung slides from TCGA-NSCLC for a binary cancer subtype  classification Lung Squamous Cell Carcinoma (LUSC) vs TCGA Lung Adenocarcinoma (LUAD) images . Classes are split between 530 LUAD and 511 LUSC examples. 

We also leveraged the Camelyon16 dataset with 400 lymph nodes sections for a 3-class classification task of metastasis detection. 240 images are labeled as normal, 80 as micro and 80 as macro. 

Finally, we use the 503 breast tissue slides from the public BRACS dataset, for a 3-class classification, identifying non-cancerous (221 slides) vs. pre-cancerous (89 slides) vs. cancerous (193 slides). 

We performed our own 10-fold stratified split for train/test splits for each dataset.

\subsection{Ablation study for encoder design}

We first conduct an ablation study on TCGA-NSCLC to help with design choices for the encoder module. We use ABMIL as base model and add components to find the best design possible. We replaced the original gated attention mechanism with a multihead dot-product attention, as described in section \ref{MIL}. The resulting ABMIL model contains 800k parameters. Adding our roformer layer increases the number of parameters to 4.3M. For fair comparison, we train larger version of ABMIL, doubling the hidden dimension size from 512 to 1024 to reach 2.1M parameters, then adding two hidden layers to reach 4.2M parameters.

The ViT layer corresponds to a transformer layer with sin-cos absolute positional encoding, similar to HIPT although they use multiple layers while we only add 1 for this ablation. The RoFormer layer is a transfomer layer with RoPE positional encoding. 

Results are reported in Table \ref{ablation}.

\input{tables/ablation}

We observe that adding a transformer layer provides a significant improvement in this classification task. However, if ViT achieved great results in the HIPT working on a small number of large patches, it fails at encoding the structure of the tissue in our setup. The very high number of small patches, tens of thousands for each image, and their inconsistent localization across slides may not be compatible with absolute position encoding. On the other hand, the relative position encoding RoPE provides an additional performance increase. This increase in performance over a global-attention pooling with similar number of parameters demonstrates that the improvement is not only due to model capacity, but to the modeling itself.

\subsection{Results}
Using the insights obtained in the last section, we evaluate the addition of a RoFormer layer to MIL models on the selected datasets.

\subsubsection{Baselines}
We apply our encoding layer to the MIL aggregators of both ABMIL\cite{ilse2018attention} and DSMIL\cite{li2021dual} and refer to the resulting models as Ro-ABMIL and Ro-DSMIL. Note that for DSMIL, we only consider the MIL aggregator and do not use the multiscale feature aggregation which is out of the scope of this work. In these experiments, we only use 1-layer encoder and keep as perspective work to scale to bigger models. We also compare our results with TransMIL that used a similar approach, as described in section \ref{transmil}, with 2.7M parameters. We computed our own TransMIL results using their public implementation.
\input{tables/results}

\subsubsection{Evaluation of performances} In Table \ref{results}, we report the AUROC on all classification tasks, macro-averaged for multi-class and with standard deviation across folds. Additional metrics are reported in Supplementary material. On all three datasets, the RoFormer layer provides a performance increase in the classification over the original ABMIL and DSMIL. TransMIL performs comparably to RoABMIL and RoDSMIL on TCGA-NSCLC, but is outperformed on both Camelyon16 and BRACS.

%% file: tables/ablation.tex
\begin{table}
\centering
\caption{Ablation study on TCGA-NSCLC}
\label{ablation}
\begin{threeparttable}
    \begin{tabular}{c c c}
    \hline
     Method & AUROC & Average Precision \\
     \hline
    ABMIL 800k & $0.86 \pm 0.06$ & $0.84 \pm 0.07$ \\
    CLAM (ABMIL 800k + instance loss) & $0.85 \pm 0.06$ & $0.83 \pm 0.08$\\
    ABMIL 2.1M & $0.88 \pm 0.06$ & $0.89 \pm 0.05$ \\
    ABMIL 4.2M & $0.88 \pm 0.06$ & $0.89 \pm 0.05$ \\
    
    Transformer layer + ABMIL   & $0.91 \pm 0.05$ & $0.91 \pm 0.04$\\
    ViT layer + ABMIL & $0.75 \pm 0.11$ & $0.77 \pm 0.09$\\
    RoFormer layer + ABMIL & $\mathbf{0.92 \pm 0.05}$& $\mathbf{0.92 \pm 0.05}$\\
    RoFormer layer + abs. position encoding + ABMIL & $0.78 \pm 0.10$ & $0.77 \pm 0.09$ \\
    \hline
    \end{tabular}

\end{threeparttable}
\end{table}

%% file: tables/results.tex
\begin{table}
\centering
\caption{Classification results}
\label{results}
\begin{threeparttable}
    \begin{tabular}{c c c c}
    \hline
     Method & TCGA-NSCLC & Camelyon16 & BRACS \\
    \hline
    ABMIL 4.2M & $0.88 \pm 0.06$ & $0.80 \pm 0.24$ & $0.77 \pm 0.09$ \\
    Ro-ABMIL & $\mathbf{0.92 \pm 0.05}$& $\mathbf{0.90 \pm 0.05}$ & $\mathbf{0.82 \pm 0.07}$\\
    DSMIL & $0.88 \pm 0.06$ & $0.74 \pm 0.23$ & $0.79 \pm 0.09$\\
    Ro-DSMIL & $0.90 \pm 0.05$ & $\mathbf{0.90 \pm 0.05}$ & $\mathbf{0.82 \pm 0.08}$\\
    TransMIL & $0.91 \pm 0.05$ & $0.84 \pm 0.09$ & $0.80 \pm 0.09$\\
    \hline
    \end{tabular}

\end{threeparttable}
\end{table}

%% file: 5_conclusion.tex
In this work, we leverage recent advances in Transformers research to design a straightforward patch encoding module as a RoFormer layer. This component addresses two of the main flaws of current approaches by modeling the inter-patches dependencies and tissue structures. The encoder is also agnostic to the downstream MIL model and we demonstrated its benefits, for both ABMIL and DSMIL, on three of the main WSI classification public datasets. We also show that the better results are due to improved modeling capabilities and not only increasing the number of parameters.

\subsubsection{Limitations and future work}
A drawback of our approach is the high number of added parameters. WSI datasets are usually small, compared to what transformers may be used to in other fields, and adding too many parameters could lead to overfitting. In particular, we presented results with only one RoFormer layer and should be careful when trying larger versions. Also, the computation overhead may be an issue for more complex approaches, in multi-modal or multi-scale scenarios for instance.

Also we stuck to a naive feature extraction scheme for simplicity, using a pretrained ResNet directly on patches, and it would be interesting to see results with features from more complex approaches, as in HIPT or DSMIL.

Finally, the datasets we used, although well known by the community, are all tumor classification on H\&E images. Applying these methods to other modalities, like IHC images, or tasks, like tumor infiltrated lymphocytes quantification, where tissue structures plays an even bigger role would be interesting.